%% file: main.tex
\documentclass[letterpaper,journal]{IEEEtran}
\usepackage{amsmath,amsfonts}
\usepackage{array}
\usepackage[caption=false,font=normalsize,labelfont=sf,textfont=sf]{subfig}
\usepackage{textcomp}
\usepackage{stfloats}
\usepackage{bm}
\usepackage{cleveref}
\usepackage{url}
\usepackage{verbatim}
\usepackage{graphicx}
\usepackage[ruled,linesnumbered]{algorithm2e}
\usepackage{float}
\usepackage{booktabs}
\usepackage{multirow}
\usepackage{pifont}
\newcommand{\cmark}{\ding{51}}
\newcommand{\xmark}{\ding{55}}
\def\BibTeX{{\rm B\kern-.05em{\sc i\kern-.025em b}\kern-.08em
    T\kern-.1667em\lower.7ex\hbox{E}\kern-.125emX}}
\usepackage{balance}
\begin{document}

\title{S$^{2}$Planner: Multi-Scale Semantic Planner for End-to-End Autonomous Driving}

\author{Zhaowei Lu, Liguo Zhou, Yujie Guo, Lei Yu, Alois Knoll~\IEEEmembership{Fellow,~IEEE,}
        % <-this % stops a space
\thanks{Zhaowei Lu, Liguo Zhou, Yujie Guo and Alois Knoll are with Chair of Robotics, Artificial Intelligence and Real-time Systems, Technical University of Munich, Garching, Germany (e-mail: zhaowei.lu@tum.de).}
\thanks{Zhaowei Lu, Liguo Zhou and Lei Yu are with Computer Science and Technology School, Huaibei Normal University, Huaibei, China (e-mail: yulei@chnu.edu.cn).}
\thanks{Liguo Zhou is the corresponding author (e-mail: liguo.zhou@tum.de).}}

\markboth{Manuscript}{}

\maketitle

\begin{abstract}
We present S$^{2}$Planner, a trajectory planner that combines three front-facing cameras with ego-motion history and the current driving command. A fine-tuned DINOv3 backbone and a Spatial Tuning Adapter produce multi-scale image features; a coarse-to-fine decoder then uses trajectory self-attention and camera-projected cross-attention to refine candidate waypoints. The contribution is the integration of ego-conditioned trajectory initialization with iterative, geometry-guided sampling of multi-scale image features, rather than a new visual backbone or attention operator. On the NAVSIM v1 non-reactive evaluation, the previously reported navtest run obtained 88.03 PDMS. Because that run was selected using navtest performance, this number is exploratory and cannot be interpreted as an unbiased test estimate. Validation-selected evaluation on unexposed data, repeated runs, and computational measurements are needed to establish generalization and efficiency.
\end{abstract}


\input{sections/1_introduction}
\input{sections/2_related_work}

\begin{figure*}
    \centering
    \includegraphics[width=\linewidth]{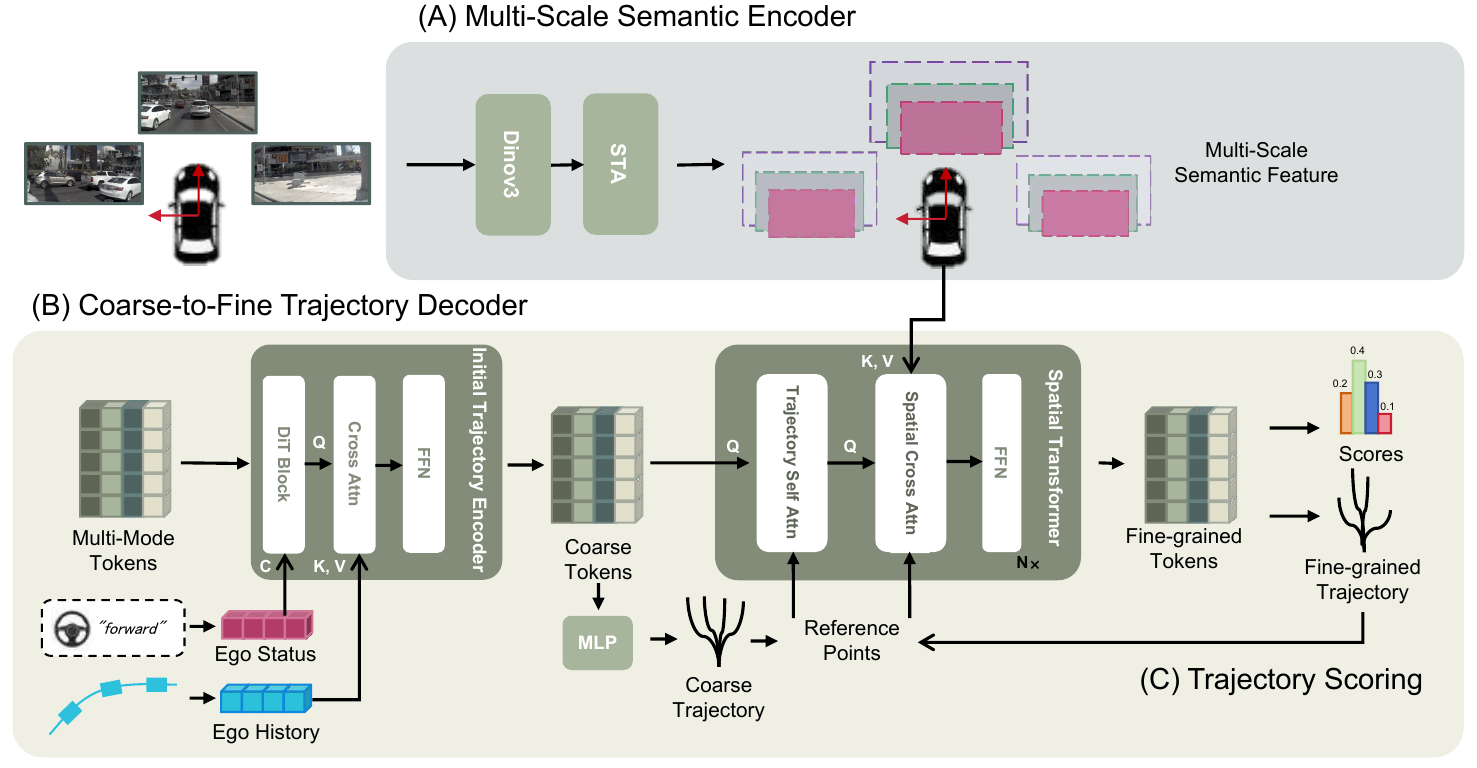}
    \caption{S$^{2}$Planner architecture. A fine-tuned DINOv3 backbone with a Spatial Tuning Adapter extracts multi-scale features from three concatenated camera images. Ego history and status condition coarse trajectory tokens and waypoints. Stacked trajectory self-attention and camera-projected cross-attention refine the candidates; the highest-scoring mode is returned.}
    \label{fig:structure_of_model}
\end{figure*}

\input{sections/3_method}
\input{sections/4_experiments}
\input{sections/5_conclusion}

%\newpage

\bibliographystyle{IEEEtran}
\bibliography{reference}

\begin{IEEEbiography}[{\includegraphics[width=1in,height=1.25in,clip,keepaspectratio]{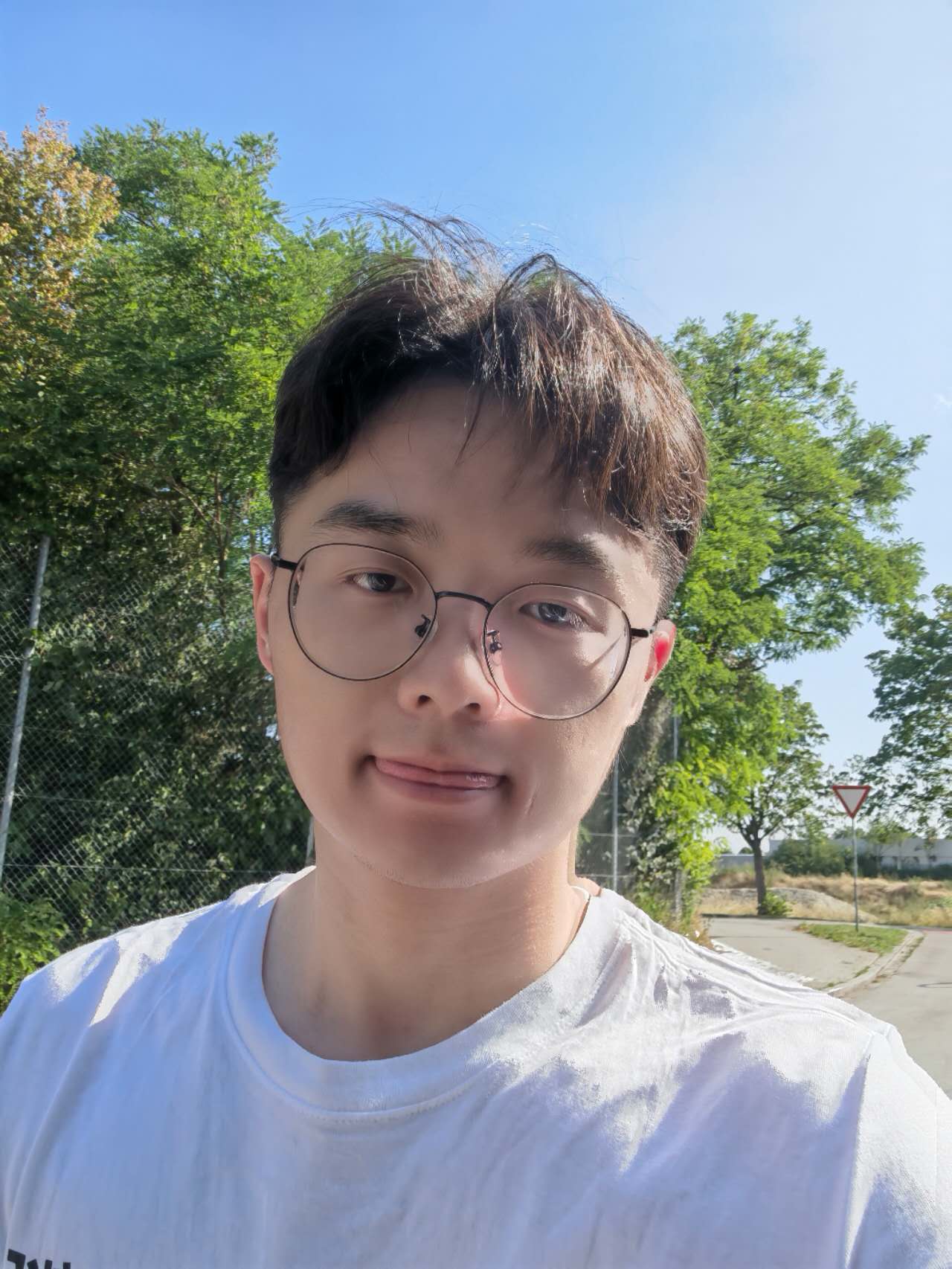}}]{Zhaowei Lu} received the Bachelor of Science degree from Tongji University and is pursuing a Master of Science degree in Robotics, Cognition, and Intelligence at the Technical University of Munich. His research interests include 3D computer vision, autonomous vehicles, generative AI, and embodied systems.
\end{IEEEbiography}

\begin{IEEEbiography}[{\includegraphics[width=1in,height=1.25in,clip,keepaspectratio]{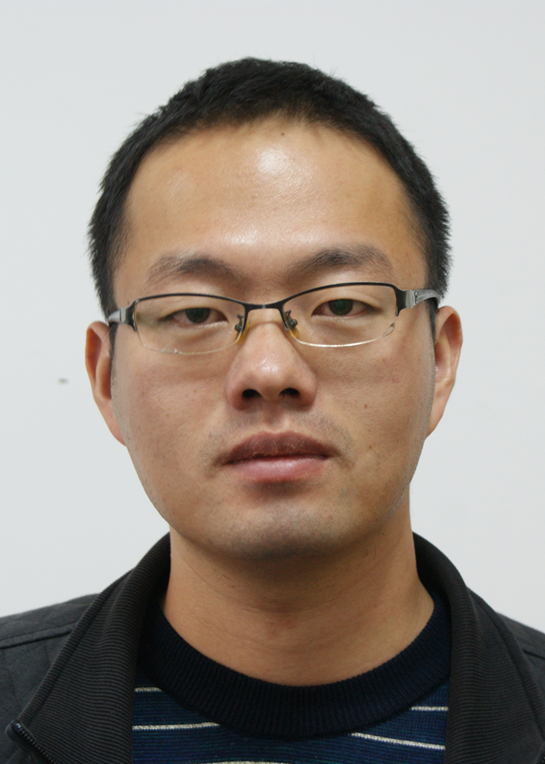}}]{Liguo Zhou} received the Bachelor of Science degree in software engineering from Suzhou University, the Master of Science degree in pattern recognition and intelligent systems from Wuhan University, and the doctoral degree in computer science from the Technical University of Munich. His research interests include computer vision, deep learning, autonomous driving, and embodied intelligence.
\end{IEEEbiography}

\begin{IEEEbiography}[{\includegraphics[width=1in,height=1.25in,clip,keepaspectratio]{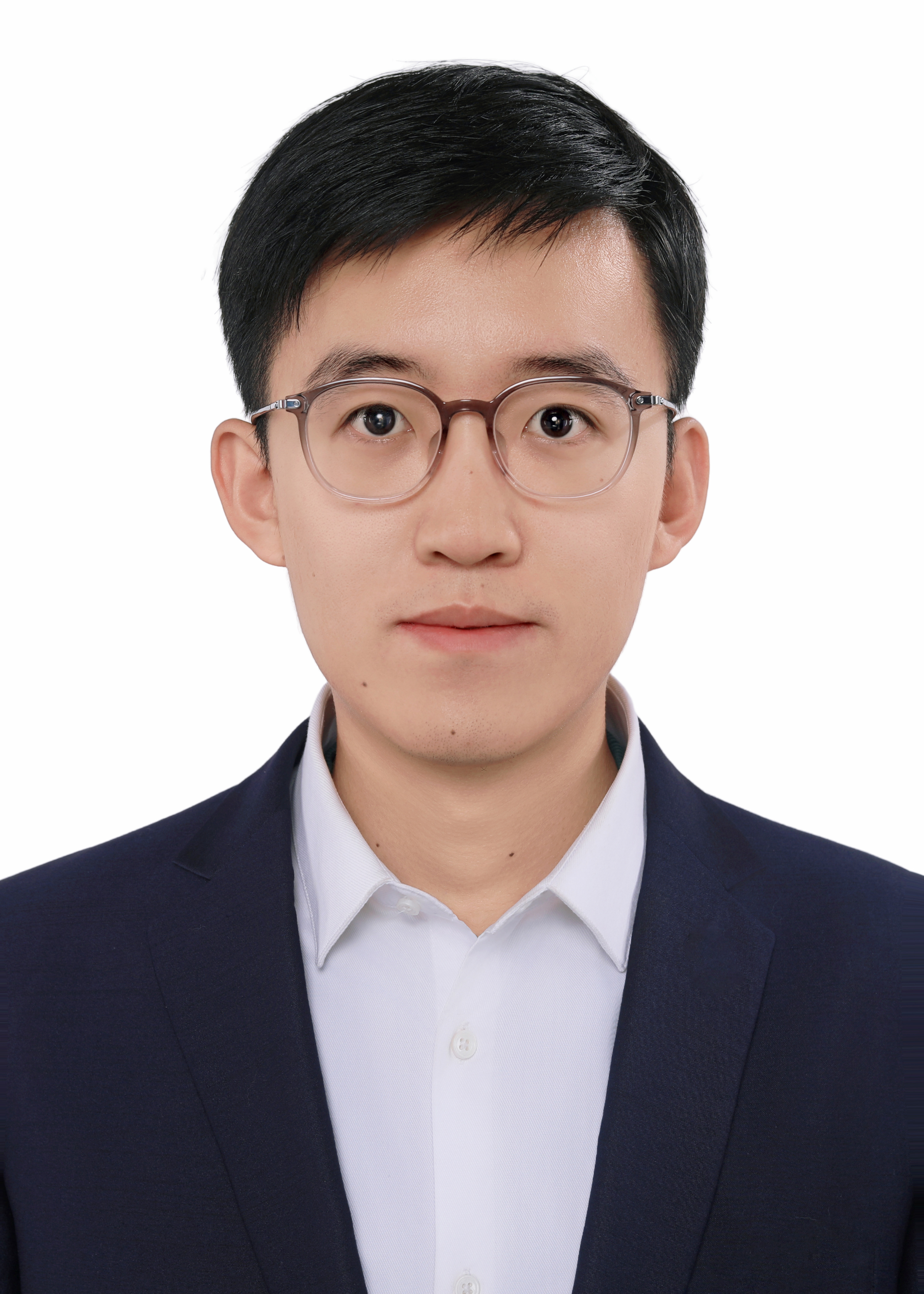}}]{Yujie Guo} received the Bachelor of Science degree from Tongji University and the Master of Science degree in Robotics, Cognition, and Intelligence from the Technical University of Munich. His research interests include deep learning, computer vision, and autonomous driving.
\end{IEEEbiography}

\begin{IEEEbiography}[{\includegraphics[width=1in,height=1.25in,clip,keepaspectratio]{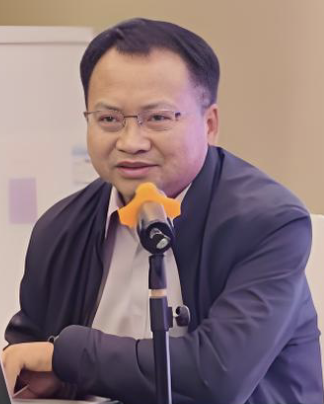}}]{Lei Yu} is a professor in the School of Computer Science and Technology at Huaibei Normal University. His research interests include artificial intelligence, information security, and security protocols.
\end{IEEEbiography}

\begin{IEEEbiography}[{\includegraphics[width=1in,height=1.25in,clip,keepaspectratio]{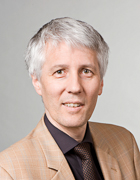}}]{Alois Knoll} (Fellow, IEEE) earned a diploma in electrical engineering from the University of Stuttgart in 1985 and a doctorate in computer science from the Technical University of Berlin in 1988. He was a professor at Bielefeld University from 1993 to 2001 and has been a professor at the Technical University of Munich since 2001. His research focuses on autonomous systems, robotics, and artificial intelligence.
\end{IEEEbiography}

\end{document}

%% file: sections/1_introduction.tex
\section{Introduction}\label{sec:intro}

\IEEEPARstart{M}{otion} planning in autonomous driving requires a model to use scene observations, the vehicle's recent motion, and a navigation command to choose a feasible future trajectory. End-to-end planners can learn this mapping directly, but differences in sensing, state inputs, evaluation protocols, and checkpoint selection complicate comparisons. NAVSIM v1~\cite{dauner2024navsim} evaluates proposed trajectories in a four-second non-reactive simulation: other actors follow recorded motion, while the ego vehicle is simulated along the proposed plan. Its Predictive Driver Model Score (PDMS) measures several aspects of plan quality, but does not establish performance in interactive traffic or a deployed closed-loop system.

Image-based planners have used BEV representations, large trajectory vocabularies, and iterative generative models~\cite{li2022bevformer,li2024hydra,liao2025diffusiondrive}. These are useful design choices, each with different modeling and computational trade-offs. We investigate whether multi-scale image features can instead be sampled directly around candidate trajectory points. Our model uses three front-facing cameras together with ego velocity, longitudinal acceleration, pose history, and a driving command. We therefore describe it as a camera-and-ego-state planner; the absence of LiDAR does not make it an image-only method.

S$^{2}$Planner combines a fine-tuned DINOv3 encoder~\cite{simeoni2025dinov3} and the Spatial Tuning Adapter from DEIMv2~\cite{huang2025real} with a trajectory decoder. Ego-conditioned tokens first generate coarse candidates. At each refinement stage, trajectory self-attention exchanges information among candidate waypoints, and spatial cross-attention samples multi-scale image features at calibrated projections of those waypoints. The candidate coordinates are updated between stages. DINOv3, the adapter, conditional transformer blocks, and deformable attention are established components; the proposed contribution is their particular trajectory-centered coupling and its evaluation in this planning setting.

An earlier navtest experiment yielded 88.03 PDMS. The checkpoint was selected by navtest score, so that result and the corresponding ablations are exploratory. We retain them to document the existing experiments while identifying the validation-selected rerun, unexposed evaluation data, and controlled comparisons required for a confirmatory result. In particular, the small difference from DiffusionDrive cannot support a superiority claim without repeated runs and uncertainty estimates.

The paper makes three contributions:

\begin{itemize}
    \item A trajectory-centered decoder that updates coarse candidates through ego-conditioned initialization and calibrated sampling of multi-scale camera features.
    \item An explicit account of how features from the concatenated camera input are associated with individual camera calibration matrices, together with the limitations introduced by attention across image seams.
    \item An exploratory NAVSIM v1 component study, with the effect of test-guided selection stated explicitly and a protocol for validation-selected follow-up evaluation.
\end{itemize}

%% file: sections/2_related_work.tex
\section{Related Works}\label{sec:related_works}

\subsection{Foundation Models for Visual Feature Learning}
Self-supervised foundation models have become a powerful means of obtaining robust visual encoders without manual labels. The DINO family exemplifies this trend: DINOv2~\cite{oquab2023dinov2} demonstrates strong transfer with self-distilled ViT backbones, and DINOv3~\cite{simeoni2025dinov3} scales training to 1.7 billion images, producing high-resolution features with strong semantic grouping. Such properties are valuable for autonomous driving, where lane markings, traffic signs, and vulnerable road users must be recognized under varied conditions.

Vision–language models such as CLIP~\cite{radford2021learning}, ALIGN~\cite{jia2021scaling}, BLIP and BLIP-2~\cite{li2022blip,li2023blip} further show that large-scale pretraining can enrich contextual understanding. Although not designed for driving, these models demonstrate how broad semantic priors can enhance downstream visuomotor tasks.

Foundation-model backbones have also been adapted to dense prediction. DEIM and DEIMv2~\cite{huang2025deim,huang2025real} integrate DINOv3 features with multi-scale matching for object detection, motivating the use of multi-scale feature hierarchies in planning. Other self-supervised approaches, including MAE~\cite{he2022masked}, BEiT~\cite{bao2021beit}, SimMIM~\cite{xie2022simmim}, and data2vec~\cite{baevski2022data2vec}, aim to learn transferable visual tokens. We study DINOv3 with a multi-scale adapter in the NAVSIM v1 non-reactive planning setting.

\subsection{Deformable Attention and Multi-Scale Transformers}
Transformers provide global context but become costly at high resolution due to dense attention. Deformable attention~\cite{xia2022vision} restricts each query to a sparse set of learned sampling locations, including across multi-scale feature maps. Deformable DETR~\cite{zhu2020deformable} and its variants~\cite{liu2022dab,meng2021conditional} use this mechanism for object detection; BEVFormer~\cite{li2022bevformer} and DETR3D~\cite{wang2022detr3d} use calibrated multi-camera image features for 3D perception.

These ideas have recently influenced sequence modeling and planning. Our approach applies stacked deformable self- and cross-attention to trajectory queries and multi-scale image features. Sparse sampling reduces the number of sampled feature locations, although end-to-end runtime must be measured before an efficiency or real-time claim can be made.

\subsection{Coarse-to-Fine and Generative Trajectory Planning}
Coarse-to-fine refinement is a long-standing strategy in vision and planning. Two-stage detectors~\cite{girshick2014rich,ren2015faster} refine coarse proposals, and trajectory forecasting methods such as ThinkTwice~\cite{jia2023think} and TrajFine~\cite{wang2024trajfine} apply similar cascades to improve waypoint quality. Diffusion-based planners such as DiffRefiner~\cite{yin2025diffrefiner} model multimodal futures through iterative denoising. S$^{2}$Planner instead produces candidate paths and refines them through a fixed stack of attention layers. The effect of this design on latency and prediction quality remains an empirical question.

\subsection{Vision-Based End-to-End Autonomous Driving}
End-to-end driving aims to map raw sensory input directly to future trajectories or controls. Early CNN-based steering models~\cite{bojarski2016end} handled only simple scenarios, motivating later works that introduced high-level commands~\cite{codevilla2018end}, multi-camera architectures, and attention mechanisms. Transformer-based planners, including cascaded decoders in ThinkTwice~\cite{jia2023think}, and BEV-based systems~\cite{li2022bevformer,huang2021bevdet,huang2022bevdet4d}, integrate scene context through cross-attention or multi-sensor fusion.

Multi-task frameworks such as YOLOP~\cite{wu2022yolop} and UniAD~\cite{hu2023planning} jointly optimize perception and planning but increase model complexity and latency. Vectorized representations, as used in VAD~\cite{jiang2023vad}, offer flexible trajectory modelling but often rely on large anchor sets or predefined motion vocabularies.

Our method combines established components in a specific way: ego-conditioned coarse trajectory tokens are updated using direct, calibrated sampling from multi-scale image features, without an explicit BEV prediction head or a fixed trajectory vocabulary. Like other NAVSIM agents, it also uses ego-state and command information. The present experiments compare selected components, but do not isolate every architectural choice or establish lower computational cost than the closest prior designs.

%% file: sections/3_method.tex
\section{Method}\label{sec:method}

\subsection{Initialization \& Tokenization}\label{sec:initialization}
We learn trajectory tokens $\bm{Q}_{\mathrm{traj}}\in\mathbb{R}^{M\times L\times D}$, where $M$ is the number of candidate modes, $L$ is the number of future waypoints, and $D$ is the token dimension. Unlike methods that initialize candidates from a fixed cluster vocabulary, this representation is learned with the planner. The coarse waypoints subsequently predicted from these tokens are still used as reference points for attention; ``anchor-free'' here refers only to the absence of a predefined trajectory vocabulary.

\subsection{Semantic-Aware Multi-Scale Scene Encoder}\label{sec:scene_encoder}
\subsubsection{Semantic Feature Extraction}\label{sec:semantic_feature_extraction}
Traditional convolution-based backbones such as ResNet architectures~\cite{he2016deep} combined with Feature Pyramid Networks (FPN)~\cite{lin2017feature} have long underpinned object detection and instance segmentation. These networks build hierarchical representations via stacked convolutions and multi-scale pyramids to mitigate the inherent single-scale nature of CNNs. Although effective in capturing local spatial structures, they often exhibit limited generalization ability and require extensive task-specific fine-tuning, increasing training cost. Their locality bias further constrains global reasoning: even with dilated convolutions or deeper designs, long-range dependencies and globally coherent semantics remain difficult to model, restricting performance in complex scene understanding tasks.

Recent advances in the DINO family have redefined feature extraction by leveraging Vision Transformers (ViT)~\cite{dosovitskiy2020image} and self-supervised learning. Through global self-attention, DINO-based models learn robust, semantically aligned, and object-centric representations that generalize well across diverse tasks with minimal fine-tuning. This cross-task transferability significantly reduces dependence on dataset-specific training and has established the DINO series as a dominant choice for modern visual backbones.

The DINO family applies self-supervised learning to vision transformers. The original DINO work~\cite{caron2021emerging} demonstrated emergent semantic structure in self-supervised ViT features; DINOv2~\cite{oquab2023dinov2} and DINOv3~\cite{simeoni2025dinov3} extended this line of work with larger-scale pretraining. These pretrained features motivate their use as a planning backbone, but their effect in this application must be established by controlled evaluation.

We use the front-left, front, and front-right cameras. After resizing each image to $H\times W$, we concatenate them in that order to form $I_{\mathrm{all}}\in\mathbb{R}^{H\times3W\times3}$. The DINOv3 backbone is initialized from DINOv3 pretraining and fine-tuned for planning. Joint self-attention lets tokens on one view access tokens from other views; it also permits attention across artificial seams between images with different camera perspectives. We do not assume that the stitched image is a single pinhole view. The downstream projection uses the calibration of each original camera. The effect of joint processing, compared with separate per-camera encoding, has not yet been isolated experimentally.

For patch size $P$ and camera dimensions divisible by $P$, the stitched input has $N=3HW/P^2$ spatial patches. We collect spatial outputs from selected DINOv3 blocks as $F^{(\ell)}=B_{\ell}(I_{\mathrm{all}})$, where $B_{\ell}$ denotes the backbone through block $\ell$. The decoder uses these spatial features, not an image-level class output. We do not specify the internal token types or positional encoding here because the exact DINOv3 variant and checkpoint have not been recovered from the original configuration.

\subsubsection{Multi-Scale Representation Module}\label{multi-scale_feature_extraction}
\begin{figure}[htb]
    \centering
    \includegraphics[width=\linewidth]{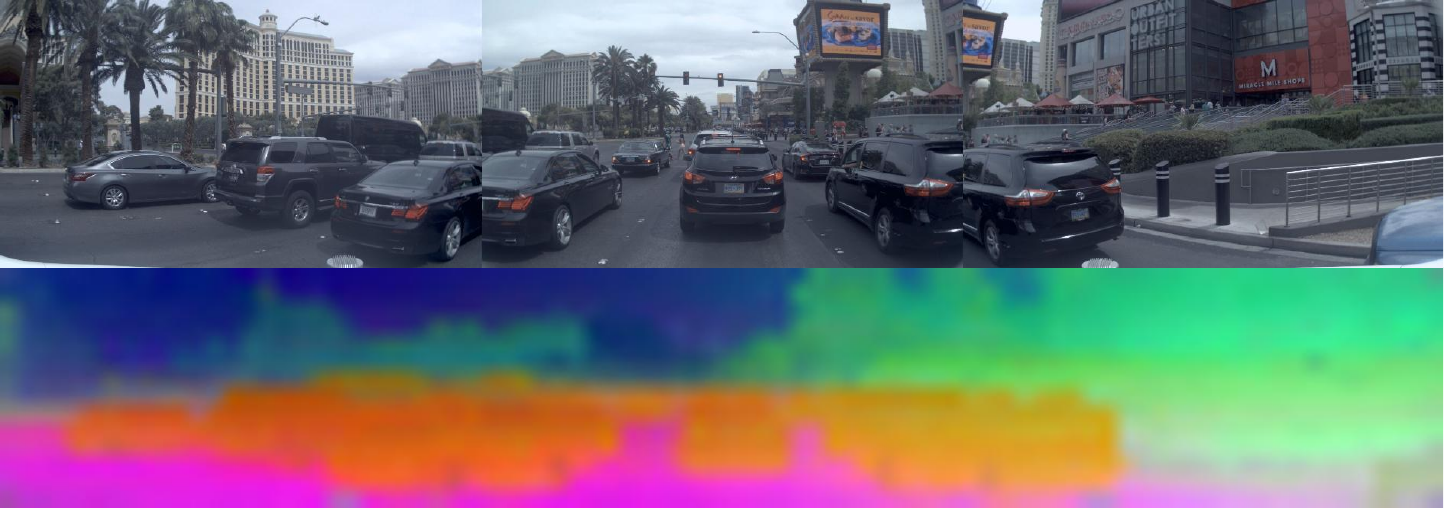}
    \caption{Concatenated front-facing images and a PCA visualization of DINOv3 patch features. This example illustrates the feature structure; a single visualization cannot establish a general loss of boundary or small-object information.}
    \label{fig:dinov3_visualization}
\end{figure}

DINOv3 patch tokens are produced on a native spatial grid. Planning can benefit from features at additional resolutions, particularly when a projected waypoint lies near a small object or boundary. \Cref{fig:dinov3_visualization} shows one example of the backbone's spatial features.

The visualization motivates multi-scale sampling, but it does not by itself measure localization or detection accuracy.

\begin{figure}[htb]
\centering
\includegraphics[width=\linewidth]{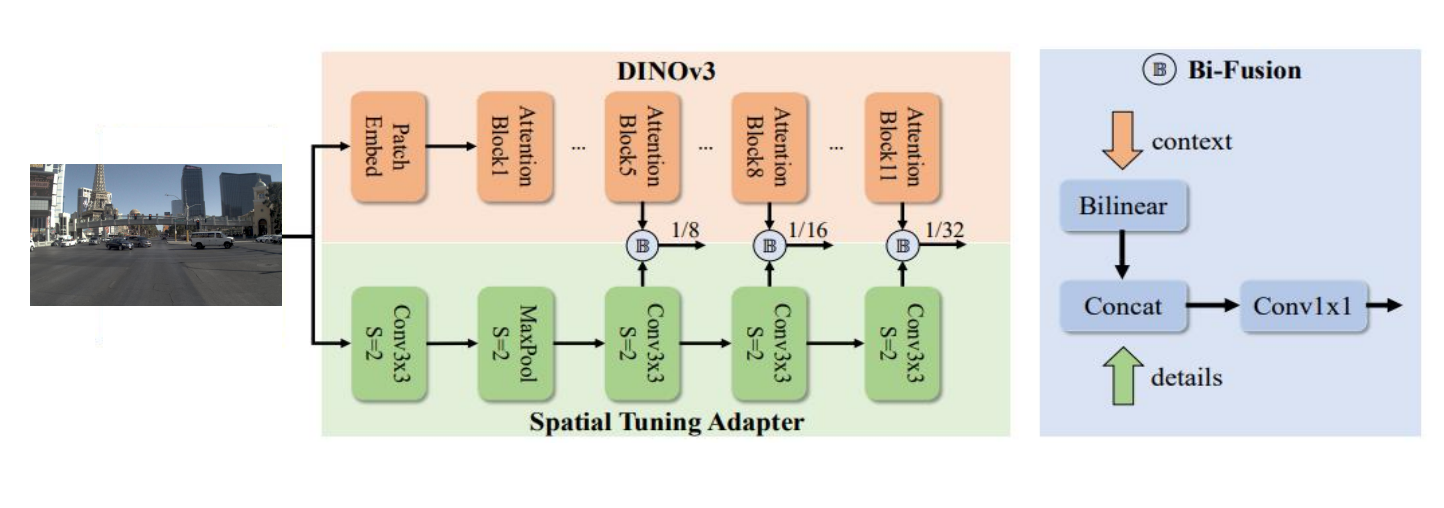}
\caption{DINOv3 and the Spatial Tuning Adapter architecture adopted from DEIMv2~\cite{huang2025real} to form multi-scale features.}
\label{fig:deimv2}
\end{figure}

We adopt the Spatial Tuning Adapter (STA) from DEIMv2~\cite{huang2025real}. It converts intermediate DINOv3 features into a multi-scale hierarchy, as illustrated in \cref{fig:deimv2}. The adapter is an existing component; its contribution to this planner is assessed through the component ablation, subject to the test-selection caveat in \cref{sec:experiments}.

Let $F^{(s)}_{\mathrm{all}}\in\mathbb{R}^{H_s\times3W_s\times D_s}$ denote a stitched feature map at scale $s$, with its width partitioned into three equal camera bands. The spatial features used by cross-attention are $F^{(s)}_i=F^{(s)}_{\mathrm{all}}[:,(i-1)W_s:iW_s,:]$ for $i\in\{1,2,3\}$, ordered front-left, front, front-right. A 3D point is projected with camera $i$'s own resized-image calibration; sampling coordinates are then normalized within $F^{(s)}_i$, not within the full $3W_s$ width. This is a spatial split of jointly encoded features, so each band can still contain information from other cameras through self-attention. Correct band boundaries require patch-aligned resizing or equivalent padding; the preprocessing implementation and feature-grid shapes must be checked against the training code before a reproducibility claim is made.

\subsection{Coarse-to-Fine Trajectory Decoder}\label{sec:trajectory_decoder}
The trajectory decoder refines an ego-conditioned coarse proposal in three stages: initial trajectory encoding, a spatial transformer, and trajectory updating and scoring (\cref{fig:structure_of_model}). Its distinguishing connection is that predicted metric waypoints determine where each refinement block samples calibrated image features.

\subsubsection{Initial Trajectory Encoding}\label{sec:initial_trajectory_encoding}
The decoder starts from the initial trajectory tokens produced in \ref{sec:initialization}. These tokens represent different candidate motion modes and are used as queries, while the ego history acts as the source of contextual information. Concretely, the ego history (e.g., past positions, headings) is first embedded by an MLP into a sequence of ego history features, which serve as keys and values in a cross-attention layer. This Initial Trajectory Encoding step injects temporal motion priors from the ego’s past behavior into each trajectory mode: tokens corresponding to different modes can attend to different segments of the ego history, enabling the model to initialize each mode with a plausible global motion trend rather than a purely data-agnostic guess.

We then apply a transformer block with scale-and-shift conditioning inspired by DiT~\cite{peebles2023scalable}. The trajectory tokens supply queries, keys, and values, while instantaneous ego speed, longitudinal acceleration, and the driving command are embedded into modulation parameters. This block does not perform diffusion sampling. The modulation gives the block access to current vehicle state; the present experiments do not isolate its contribution.

A feed-forward network with a residual connection updates the tokens, and an MLP predicts coarse future waypoints. This produces $\bm T_{\mathrm{coarse}}\in\mathbb{R}^{M\times L\times3}$ and $\bm Q_{\mathrm{traj,coarse}}\in\mathbb{R}^{M\times L\times D}$. At this point the candidates are conditioned on ego history and status; image information enters later through spatial cross-attention. Algorithm~\ref{alg:init_traj_encoding} summarizes this initialization.

\begin{algorithm}[t]
    \LinesNumberedHidden
    \small
    \caption{Initial Trajectory Encoding with Conditional DiT Block}
    \label{alg:init_traj_encoding}
    
    \KwIn{Trajectory tokens $Q_{\text{traj}}$}
    \KwIn{Ego history $H_{\text{ego}}$, Status $s_{\text{ego}}$}
    \KwOut{$T_{\text{coarse}}$, $Q_{\text{traj-coarse}}$}
    
    $Q_{\text{hist}} \leftarrow \text{HistEncoder}(H_{\text{ego}})$
    \tcp*{Encode ego history}
    
    $Q_{\text{coarse}} \leftarrow \text{CrossAttn}(Q_{\text{traj}}, Q_{\text{hist}})$\;
    \tcp*{Inject priors}
    
    $(\gamma, \beta) \leftarrow \text{CondMLP}(s_{\text{ego}})$
    \tcp*{Compute modulation}
    
    $\hat{Q} \leftarrow \text{LayerNorm}(Q_{\text{coarse}})$\;
    
    $\hat{Q} \leftarrow \gamma \odot \hat{Q} + \beta$;
    \tcp*{Apply condition}
    
    $Q_{\text{DiT}} \leftarrow Q_{\text{coarse}} + \text{TransformerBlock}(\hat{Q})$\;
    \tcp*{Self Attn}
    
    $Q_{\text{traj-coarse}} \leftarrow \text{FFN}(Q_{\text{DiT}})+Q_{\text{DiT}}$
    \tcp*{Final projection}
    
    $T_{\text{coarse}} \leftarrow \text{MLP}(Q_{\text{traj-coarse}})$
    \tcp*{Output coarse traj}
    
    \Return $T_{\text{coarse}}$, $Q_{\text{traj-coarse}}$\;

\end{algorithm}

\subsubsection{Spatial Transformer}\label{sec:spatial_transformer}
The spatial transformer contains trajectory self-attention (TSA) and spatial cross-attention (SCA). TSA exchanges information among candidate trajectory tokens; SCA samples the multi-scale camera features around calibrated projections of the current trajectory points.

Both attention operators adapt deformable sampling from prior work~\cite{li2022bevformer,zhu2020deformable}. We use the current predicted waypoints as reference locations rather than a fixed library of trajectory anchors. Sparse sampling reduces attention locations, but whole-model speed and memory have not been measured.

\paragraph{Trajectory Self Attention}
In \textbf{Trajectory Self Attention} (TSA), coarse trajectory points $\bm{T}_{\mathrm{coarse}}\in\mathbb{R}^{M\times L\times3}$ provide reference coordinates in the current ego frame. The manuscript specifies a forward region $x\in[0,32]$ m and lateral region $y\in[-32,32]$ m, with positive $x$ forward and positive $y$ to the left. The center waypoints and four ego-footprint corners are normalized to this region as proposed reference locations~\cite{guo2025ipad}. The number and distribution of any additional sampled locations remain to be verified against the implementation.

\begin{figure}[htb]
\centering
\includegraphics[width=0.45\linewidth]{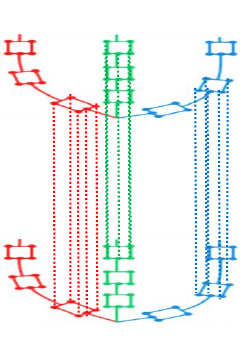}
\caption{Visualization of the anchor-based feature sampling. The interaction of the queries in the trajectory feature space is activated while the dense attention computation is prevented. For the sake of simplicity, the offset is not presented here.}
\label{fig:tsa_visualization}
\end{figure}

For TSA, queries and values originate from the coarse trajectory tokens $\bm Q_{\mathrm{traj,coarse}}$ generated in \cref{sec:initial_trajectory_encoding}. We summarize the intended operation abstractly as

\begin{equation}
\text{TSA}(\mathit{Q}_{m}^{p}, \bm{V}_{TSA})=\sum\text{DeformAttn}(\mathit{Q}_{m}^{p}, \mathit{p}, \bm{V}_{TSA})
\label{eq:TSA}
\end{equation}
where $Q_m^p$ is the query for mode $m$ at waypoint $p$. Learned offsets can shift reference locations and permit information exchange between candidate modes. The $M\times L\times D$ token tensor is indexed by mode and time, however, not by a regular $(x,y)$ image grid. Therefore, the physical meaning of bilinear sampling in this tensor, its spatial layout, and its offset parameterization cannot be established from the manuscript alone and must be specified from the implementation. \Cref{fig:tsa_visualization} is a conceptual illustration of reference-point sampling, not evidence of a particular interpolation rule.

\paragraph{Spatial Cross Attention}
The updated trajectory feature tokens $\bm{Q}_{traj-TSA}\in\mathbb{R}^{M \times L \times D}$ from Trajectory Self Attention module are fed into the \textbf{Spatial Cross Attention} (SCA) layer, where they act as queries, while the multi-scale semantic features from the visual backbone serve as keys and values. Instead of applying standard multi-head attention directly over all views and all feature tokens with different scale, which leads to prohibitive computational and memory costs, the SCA implements the deformable attention to 3D space. In particular, each trajectory query token $\bm{q}$ attends only to a small set of spatially relevant locations across the camera views, rather than densely interacting with all image features. Also, since the deformable attention was originally proposed for purely 2D perception tasks, where queries and sampled features lie in the same image plane, several modifications are necessary when extending it to 3D scenes and trajectory representations.

The coarse trajectory point $p=(x,y)$ remains in the current ego vehicle's metric frame. Following the pillar-style construction in~\cite{li2022bevformer}, $N_{\mathrm{ref}}$ heights $z_j$ produce 3D reference points. Each point is projected independently into each calibrated camera $i$:

\begin{align}
\tilde{\bm u}_{ij} &= P_i[x,y,z_j,1]^{\mathsf T}, \\
\mathcal{P}_i(p,z_j) &=
 (\tilde u_{ij,1}/\tilde u_{ij,3},
  \tilde u_{ij,2}/\tilde u_{ij,3})
\end{align}
where $P_i\in\mathbb{R}^{3\times4}$ maps ego-frame 3D points to camera $i$'s resized pixel grid. A projection is valid only if its depth $\tilde u_{ij,3}$ is positive and its pixel coordinate is inside that camera's $W\times H$ image. For scale $s$, valid pixel coordinates are transformed to the local coordinate system of $F_i^{(s)}$ before deformable sampling. In particular, the front-camera coordinate is never offset by the width of the front-left image during calibration projection; the offset is used only to extract its band from the stitched feature tensor.

\begin{figure}[H]
\centering
\includegraphics[width=\linewidth]{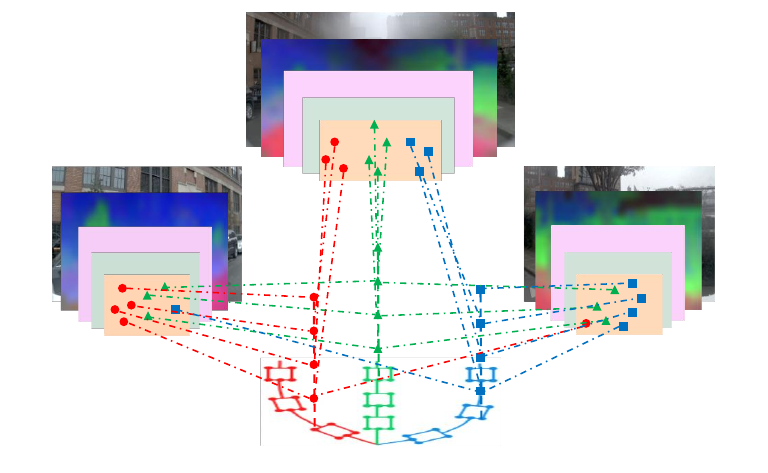}
\caption{Illustration of the principle of Spatial Cross Attention. The reference anchors are directly derived from the actual trajectory coordinates, making sure that the metric consistency is preserved. For each trajectory query, only the feature from region where the projected 2D point locates inside the image plane would be seen as valid. The new trajectory feature is computed based on a weighted aggregation of the valid image features}
\label{fig:sca_vis}
\end{figure}

Finally, all sampled features would be aggregated based on a set of softmax-normalized attention weights generated by a MLP with $\bm{Q}_{traj-TSA}$ as input. The overall process could be summarized as:

\begin{equation}
\begin{aligned}
&\text{SCA}(\bm{Q}_{m}^{p}, \bm{F}) = \\
&\qquad\frac{1}{|\mathcal{N}_{hit}|}
\sum_{i\in\mathcal{N}_{hit}}\sum_{j=1}^{N_{\mathrm{ref}}}
\sum_s\text{DeformAttn}\big(\bm{Q}_{m}^{p}, \mathcal{P}_i(p,z_j), \bm{F}_{i}^{(s)}\big)
\end{aligned}
\label{eq:SCA}
\end{equation}
Here $i$, $j$, and $s$ index the camera, pillar height, and feature scale. $\mathcal{N}_{hit}$ is the set of cameras with valid projections, and the expression is evaluated only when that set is nonempty; the implementation's fallback for an empty set must be verified. $\bm Q_m^p$ is the query for mode $m$ at waypoint $p$. As illustrated in \cref{fig:sca_vis}, invalid projections should be masked before attention weights are normalized; the exact masking behavior requires code verification. The per-view tensors remain jointly encoded because DINOv3 attended across the concatenated input; a separate-view encoder is needed to measure whether cross-seam attention helps or harms.

\subsubsection{Trajectory Updating \& Scoring}\label{sec:trajectory_updating_scoring}
The Trajectory Self Attention module and the Spatial Cross Attention module together form the core of a \textbf{Spatial Transformer} block. This block is stacked $K$ times to obtain a hierarchical refinement architecture for trajectory prediction. After each iteration, the refined trajectory tokens are passed through an MLP decoder to predict trajectory coordinates directly. These trajectory coordinates are the updated  trajectory points, yielding progressively more accurate trajectory estimates. The updated points then serve as new reference points for the next refinement stage. The final fine-grained trajectory tokens would be also processed by another MLP decoder, to get scores for each of the mode of trajectories. The whole process could be summarized in following:

\begin{align}
\bm{Q}^{k} &= SF_{k}(\bm{Q}^{k-1}), \\
\bm{T}^{k} &= MLP(\bm{Q}^{k}), \\
\bm{S}_{pred} &= MLP(\bm{Q}^{K}).
\end{align}
where $k = 1,\dots,K$ indexes the refinement stage.  $\bm{Q}^{k}$ illustrates the output trajectory queries of $k$-th Spatial Transformer layer $SF_{k}$, $\bm{T}^{k}$ denotes the corresponding trajectory predictions decoded from $\bm{Q}^{k}$, and $\bm{S}_{pred}$ denotes the final set of scores associated with the trajectories derived from the last-stage tokens $\bm{Q}^{K}$.

The decoder thus uses updated trajectory coordinates to choose the image sampling locations at the next stage. Its effect on final planning scores is examined through the exploratory component ablation in \cref{sec:ablation_study}.

\subsection{Loss}\label{sec:loss}
The overall training objective is composed of two main components: a trajectory regression loss and a score-based classification loss. The trajectory loss supervises the predicted future motion at multiple refinement stages, while the score loss supervises the mode probabilities associated with the final trajectory set. The first part of the trajectory loss is the intermediate trajectory loss applied to multiple stages of the refinement pipeline. Inspired by ~\cite{guo2025ipad},  We apply a simple Minimum over N (MoN) loss~\cite{gupta2018social}, which is defined as:

\begin{equation}
\mathcal{L}_{intermediate} = \sum_{k=0}^{K}\alpha^{K-k}\min\limits_{m\in[0,...,M-1]}\|\hat{\boldsymbol{T}}_{k}^{m}-\boldsymbol{T}_{gt}\|
\label{eq: Intermediate trajectory loss}
\end{equation}
As written, $k=0$ denotes the coarse prediction and $k=1,\ldots,K$ denotes the $K$ spatial-transformer outputs. Thus the sum includes the final stage, which also receives the separate final loss below. The intended supervision stages and decay factor $\alpha$ must be checked against the training code. Here $\hat{\bm T}_k^m$ is mode $m$ at stage $k$, and $\bm T_{gt}\in\mathbb{R}^{L\times3}$ is the expert trajectory.

The second part of the trajectory loss is applied at the final refinement stage and directly supervises the multi-modal set of trajectories and their associated scores. For the final layer output $\hat{\bm{T}}_{final}\in\mathbb{R}^{M \times L \times 3}$ and the predicted scores $\bm{S}_{pred}\in\mathbb{R}^{M}$, we construct a one-hot mode label $\bm{S}_{label}\in\mathbb{R}^{M}$ as follows:
\[
S_{label}^{m} =
\begin{cases}
1, & \text{if } m = m^*,\\
0, & \text{otherwise}.
\end{cases}
\]
Specifically, the trajectory points $\hat{\bm{T}}_{final}^{m^*}\in\mathbb{R}^{L \times 3}$
that is closest to the expert trajectory points $\bm{T}_{gt}\in\mathbb{R}^{L \times 3}$ (in terms of the $\ell_1$ distance measure) is identified and labeled as positive, while all other modes are labeled as negative.

Given the predicted scores $\bm{S}_{pred}$ and the label vector
$\bm{S}_{label}$, two particular trajectories are then extracted from the
multi-modal set. First, a predicted-best trajectory points $\hat{\bm{T}}_{pred}\in\mathbb{R}^{L \times 3}$ is obtained by selecting the trajectory mode with the highest predicted score:
\begin{equation}
\hat{\bm{T}}_{pred} = 
\hat{\bm{T}}_{final}^{\arg\max_m \bm{S}_{pred}^{m}}
\end{equation}
Second, a label-best trajectory points $\hat{\bm{T}}_{label}\in\mathbb{R}^{L \times 3}$ is obtained by selecting the trajectory mode with the highest label value, which is equivalent to the distance-based best mode:
\begin{equation}
\hat{\bm{T}}_{label} = 
\hat{\bm{T}}_{final}^{\arg\max_m S_{label}^{m}} = 
\hat{\bm{T}}_{final}^{m^*}
\end{equation}

For the score supervision, the score loss is formulated as:

\begin{equation}
\mathcal{L}_{label} = \mathcal{L}_{BCE}(\sigma(\bm{S}_{pred}), \bm{S}_{label})
\label{eq:Label_loss}
\end{equation}
where $\sigma$ denotes the sigmoid function applied element-wise to the predicted scores. The final loss could then be defined as:

\begin{equation}
\mathcal{L}_{f} = \frac{\mathcal{L}_{pred-gt}+\mathcal{L}_{label-gt}+\mathcal{L}_{pred-label}}{3}+\lambda\mathcal{L}_{label}
\end{equation}
where each $\mathcal{L}_{traj_1-traj_2}$ denotes an $\ell_1$ loss between two sets of trajectory points, and $\lambda$ is a weighting factor that balances the trajectory regression terms and the score supervision term $\mathcal{L}_{label}$. By averaging $\mathcal{L}_{pred-gt}$, $\mathcal{L}_{label-gt}$ and $\mathcal{L}_{pred-label}$, the final-stage trajectory loss ensures that the model learns to produce at least one accurate trajectory in the multi-modal set, to assign high scores to accurate trajectories, and to maintain consistency between the scoring mechanism and the regression quality. The total loss is $\mathcal{L}_{total} = \mathcal{L}_{f} + \mathcal{L}_{intermediate}$.

%% file: sections/4_experiments.tex
\section{Experiments}\label{sec:experiments}

\subsection{Benchmark and Evaluation Protocol}\label{sec:dataset}
We study NAVSIM v1~\cite{dauner2024navsim}, whose official filtered splits are navtrain and navtest; there is no standard navval filter in the v1.1 split documentation. A validation set for model selection must therefore be constructed from navtrain with driving logs kept disjoint between training and validation, and its scene identifiers published. NAVSIM evaluates an agent-proposed trajectory in a four-second \emph{non-reactive} simulation: background agents follow recorded future motion, while an LQR controller tracks the proposed ego trajectory. The planner is not queried repeatedly in an interactive rollout. PDMS therefore measures the proposed trajectory under this fixed-future evaluation, not long-horizon closed-loop behavior with agents reacting to the ego vehicle.

The NAVSIM v1.1 evaluator combines no-at-fault collision (NC) and drivable-area compliance (DAC) as multiplicative factors with ego progress (EP), time-to-collision (TTC), and comfort (C) as weighted terms~\cite{dauner2024navsim}:
\begin{equation}
\mathrm{PDMS}=\mathrm{NC}\,\mathrm{DAC}
\left(\frac{5\mathrm{EP}+5\mathrm{TTC}+2\mathrm{C}}{12}\right).
\label{eq:PDMS}
\end{equation}
The reference implementation is the NAVSIM v1.1 scorer at
\url{https://github.com/autonomousvision/navsim/blob/v1.1/navsim/planning/simulation/planner/pdm_planner/scoring/pdm_scorer.py}.
The official per-scenario scores are aggregated by the evaluator. Equation~\eqref{eq:PDMS} must not be applied to rounded, dataset-averaged columns to reconstruct dataset PDMS.

\subsection{Implementation Details and Reproducibility}\label{sec:implementation_details}
\subsubsection{Inputs and Optimization}
S$^{2}$Planner uses front-left, front, and front-right RGB cameras and four categories of non-image input: current ego velocity, longitudinal acceleration, recent pose history in the ego frame, and a driving command. Thus the camera entry in the sensor column of \cref{tab:quantitative_comparison} indicates the absence of LiDAR; it does not indicate image-only input. Images are resized and calibration matrices must be adjusted to the resized pixels. The backbone is initialized from DINOv3 pretrained weights and fine-tuned; it is not initialized from an ImageNet-supervised classification checkpoint. The trajectory decoder is trained from scratch.

The available manuscript records Adam with $\beta_1=0.9$, $\beta_2=0.999$, initial learning rate $10^{-4}$, polynomial decay to $10^{-5}$, weight decay $10^{-2}$, gradient clipping at norm $1$, global mini-batch size $128$, mixed precision, and eight NVIDIA A100 GPUs. The archived text does not specify the DINOv3 variant, per-camera image resolution, patch handling at camera seams, number of trajectory modes, waypoint interval, deformable-attention sampling count, exact loss weights, training epochs or steps, or random seeds. These values must be recovered from the experiment configuration before the study can be reproduced. The notation $M$, $L$, $N_{\mathrm{ref}}$, $K$, $\alpha$, and $\lambda$ in \cref{sec:method} does not substitute for configuration values. The main architecture figure also depicts a different order of history attention and conditional transformer processing from Algorithm~\ref{alg:init_traj_encoding}; the executed order must be checked in code.

\subsubsection{Checkpoint Selection and Result Status}
In the original experiments, the checkpoint was selected by the highest \emph{navtest} PDMS. This exposes the test set to model selection and can bias both the main result and navtest ablations upward. All S$^{2}$Planner values retained below are consequently \emph{exploratory, test-selected observations}, including the reported 88.03 PDMS. They are not independent estimates of generalization. A corrected development protocol would select the checkpoint and all hyperparameters on a log-disjoint validation subset of navtrain, freeze the configuration, and then evaluate navtest once per selected checkpoint. Because navtest has already influenced this project, such a rerun on the same split cannot restore its status as untouched evidence. An independent generalization claim requires a genuinely unexposed evaluation set. The ablations also need to be rerun under the corrected selection protocol.

\subsubsection{Required Statistical and Computational Reporting}
The original records provide one score per model and no scenario-level score files, repeated-seed results, parameter counts, FLOPs, peak inference memory, batch-one latency, or throughput. Accordingly, this paper does not infer statistical significance, low latency, or real-time suitability from the existing table. A complete comparison should report multiple training seeds as mean and standard deviation, a paired scenario-level analysis on the same navtest scenes, and timing on the same hardware with synchronized inference and a specified precision, input resolution, batch size, and warm-up. Baseline resource figures should be measured under the same conditions or clearly identified as numbers reported by their authors.

\subsection{Exploratory Quantitative Results}\label{sec:quantitative_results}
\begin{table*}[t]
\centering
\caption{Previously reported NAVSIM v1 navtest results (percent). Our row is test-selected and exploratory; the other rows are literature results reproduced from the original comparison, with their selection protocols and non-image inputs not audited here. Camera describes the sensor category only. NC, DAC, TTC, comfort (C), and EP are dataset summaries; dataset PDMS is aggregated from scenario scores. These rows do not establish a statistically significant ranking.}
\label{tab:quantitative_comparison}
\renewcommand{\arraystretch}{1.15}
\begin{tabular*}{\textwidth}{l @{\extracolsep{\fill}} c c c c c c c c}
\toprule
\textbf{Method} & \textbf{Sensors} & \textbf{Anchors} & NC & DAC & TTC & C & EP & PDMS \\
\midrule
Transfuser~\cite{chitta2022transfuser} & Camera + LiDAR & 0 & 97.7 & 92.8 & 92.8 & 100 & 79.2 & 84.0 \\
DRAMA~\cite{yuan2024drama} & Camera + LiDAR & 0 & 98.0 & 93.1 & 94.8 & 100 & 80.1 & 85.5 \\
VADv2-V$_{8192}$~\cite{chen2024vadv2} & Camera + LiDAR & 8192 & 97.2 & 89.1 & 91.6 & 100 & 76.0 & 80.9 \\
Hydra-MDP-V$_{8192}$~\cite{li2024hydra} & Camera + LiDAR & 8192 & 97.9 & 91.7 & 92.9 & 100 & 77.6 & 83.0 \\
Hydra-MDP-V$_{8192}$-W-EP~\cite{li2024hydra} & Camera + LiDAR & 8192 & 98.3 & 96.0 & 94.6 & 100 & 78.7 & 86.5 \\
DiffusionDrive~\cite{liao2025diffusiondrive} & Camera + LiDAR & 20 & 98.2 & 96.2 & 94.7 & 100 & 82.2 & 88.1 \\
\midrule
UniAD~\cite{hu2023planning} & Camera & 0 & 97.8 & 91.9 & 92.9 & 100 & 78.8 & 83.4 \\
PARA-Drive~\cite{weng2024drive} & Camera & 0 & 97.9 & 92.4 & 93.0 & 99.8 & 79.3 & 84.0 \\
LTF~\cite{chitta2022transfuser} & Camera & 0 & 97.4 & 92.8 & 92.4 & 100 & 79.0 & 83.8 \\
S$^{2}$Planner (test-selected) & Camera + ego state & 0 & 98.2 & 96.5 & 94.3 & 100 & 82.2 & 88.03 \\
\bottomrule
\end{tabular*}
\end{table*}

The previously reported S$^{2}$Planner PDMS of 88.03 is close to DiffusionDrive's reported 88.1. Their sensors, state channels, compute budgets, and selection procedures have not been matched, and the S$^{2}$Planner score is test-selected. We therefore cannot conclude that either method is better from this comparison. An ego-state-only baseline and a camera-plus-state baseline would help quantify the contribution of visual information beyond the state signals.

\subsection{Qualitative Examples}\label{sec:qualitative_results}
\begin{figure*}[t]
    \centering
    \includegraphics[width=\textwidth]{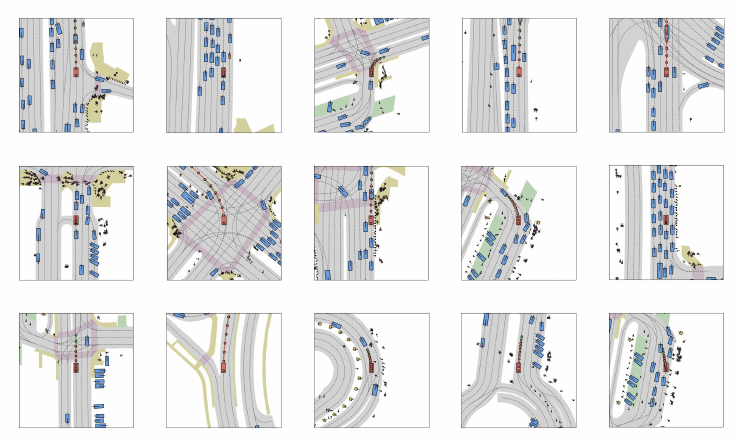}
    \caption{Selected navtest examples. Green shows the displayed expert path and red the predicted path. The displayed paths may cover different time horizons; their apparent lengths should not be interpreted as a planning error or a progress score. These static overlays do not show how other agents would react to the plan.}
    \label{fig:visualizations}
\end{figure*}

\Cref{fig:visualizations} displays selected predicted trajectories against expert paths. In these images, several predictions visually follow the road corridor and avoid visible obstacles. Such overlays are illustrative; they do not establish collision freedom, anticipation of another agent's future action, comfort, or human-like behavior. Those properties require quantitative scenario-level metrics or reactive evaluation. The plotted expert and predicted paths also need matched temporal horizons for a direct geometric comparison.

\subsection{Exploratory Ablations}\label{sec:ablation_study}
The earlier component experiments used navtest in a test-guided development cycle. Their differences are descriptive and need replication with checkpoints selected on a log-disjoint validation subset of navtrain. The results in \cref{tab:ablation_check} do not identify the causal effect of one component unless all other training choices, input channels, and compute budgets were held fixed.

\begin{table}[t]
\centering
\caption{Previously reported component ablations on navtest. Values are exploratory and test-selected; repeated seeds and a validation-selected rerun are pending.}
\label{tab:ablation_check}
\resizebox{\columnwidth}{!}{
\begin{tabular}{l c c c c}
\toprule
\textbf{Variant} & \textbf{DINOv3} & \textbf{STA} & \textbf{Deform.} & \textbf{PDMS} \\
\midrule
A0: ResNet & \xmark & \xmark & \xmark & 72.4 \\
A1: DINOv3 & \cmark & \xmark & \xmark & 84.9 \\
A2: DINOv3 + STA & \cmark & \cmark & \xmark & 86.3 \\
A3: full model & \cmark & \cmark & \cmark & 88.0 \\
\bottomrule
\end{tabular}}
\end{table}

Within the reported runs, A1 exceeds A0 by 12.5 points, A2 exceeds A1 by 1.4 points, and A3 exceeds A2 by 1.7 points at the displayed precision. Because these comparisons were developed on navtest and have no uncertainty estimates, they cannot establish which component reliably improves held-out performance. A controlled architecture study should additionally compare separate versus concatenated camera encoding, matched state inputs, and the closest trajectory-refinement decoder under equal training and compute budgets.

\begin{table}[t]
\centering
\caption{Previously reported BEV auxiliary-head comparison on navtest. This tests one BEV-head design and loss setting only; values are exploratory.}
\label{tab:ablation_bev}
\begin{tabular}{l c c}
\toprule
\textbf{Variant} & \textbf{BEV head} & \textbf{PDMS} \\
\midrule
A1: direct regression & \xmark & 84.9 \\
A1.1: A1 + BEV auxiliary head & \cmark & 83.5 \\
\bottomrule
\end{tabular}
\end{table}

The single BEV comparison in \cref{tab:ablation_bev} shows a lower score for this particular auxiliary head. It does not show that BEV representations are generally unnecessary: head capacity, loss weighting, gradient interference, and training variance could all affect the result. A stronger test would vary the BEV loss weight and head design under the same validation-selected protocol.

%% file: sections/5_conclusion.tex
\section{Conclusions and Future Work}\label{sec:conclusion}

S$^{2}$Planner combines a fine-tuned DINOv3 and Spatial Tuning Adapter with an ego-conditioned, coarse-to-fine trajectory decoder. Its specific design samples multi-scale camera features at calibrated projections of candidate waypoints and updates the waypoints between refinement stages. The model uses three RGB cameras, ego velocity, acceleration, pose history, and a driving command.

The previously reported NAVSIM v1 result of 88.03 PDMS was obtained after selecting a checkpoint using navtest. It is therefore exploratory, and the available single-run ablations and qualitative examples cannot establish generalization, a statistically meaningful advantage over close baselines, real-time performance, or interactive driving competence. The BEV experiment applies to one auxiliary head and does not support a general conclusion about BEV representations.

The next experimental step is to create a log-disjoint validation subset from navtrain, select all settings there, and freeze the configuration before further evaluation. Since this project has already used navtest for development, a fresh independent generalization claim requires an unexposed evaluation set. Repeated seeds, paired scenario-level analyses, an ego-state-only control, a separate-camera-encoding control, and matched comparisons with the closest refinement methods are needed. Parameter count, FLOPs, latency, memory, and throughput should be measured under a specified hardware and software setup. Finally, interactive simulation or real-world testing is necessary to evaluate reactions to other agents and behavior beyond the four-second non-reactive NAVSIM v1 horizon.

%% file: reference.bib
@inproceedings{liao2025diffusiondrive,
  title={Diffusiondrive: Truncated diffusion model for end-to-end autonomous driving},
  author={Liao, Bencheng and Chen, Shaoyu and Yin, Haoran and Jiang, Bo and Wang, Cheng and Yan, Sixu and Zhang, Xinbang and Li, Xiangyu and Zhang, Ying and Zhang, Qian and others},
  booktitle={Proceedings of the Computer Vision and Pattern Recognition Conference},
  pages={12037--12047},
  year={2025}
}

@inproceedings{he2016deep,
  title={Deep residual learning for image recognition},
  author={He, Kaiming and Zhang, Xiangyu and Ren, Shaoqing and Sun, Jian},
  booktitle={Proceedings of the IEEE conference on computer vision and pattern recognition},
  pages={770--778},
  year={2016}
}

@inproceedings{lin2017feature,
  title={Feature pyramid networks for object detection},
  author={Lin, Tsung-Yi and Doll{\'a}r, Piotr and Girshick, Ross and He, Kaiming and Hariharan, Bharath and Belongie, Serge},
  booktitle={Proceedings of the IEEE conference on computer vision and pattern recognition},
  pages={2117--2125},
  year={2017}
}

@inproceedings{caron2021emerging,
  title={Emerging Properties in Self-Supervised Vision Transformers},
  author={Caron, Mathilde and Touvron, Hugo and Misra, Ishan and J\'{e}gou, Herv\'{e} and Mairal, Julien and Bojanowski, Piotr and Joulin, Armand},
  booktitle={Proceedings of the IEEE/CVF International Conference on Computer Vision},
  pages={9650--9660},
  year={2021},
  doi={10.1109/ICCV48922.2021.00951}
}

@article{oquab2023dinov2,
  title={Dinov2: Learning robust visual features without supervision},
  author={Oquab, Maxime and Darcet, Timoth{\'e}e and Moutakanni, Th{\'e}o and Vo, Huy and Szafraniec, Marc and Khalidov, Vasil and Fernandez, Pierre and Haziza, Daniel and Massa, Francisco and El-Nouby, Alaaeldin and others},
  journal={arXiv preprint arXiv:2304.07193},
  year={2023}
}

@article{simeoni2025dinov3,
  title={Dinov3},
  author={Sim{\'e}oni, Oriane and Vo, Huy V and Seitzer, Maximilian and Baldassarre, Federico and Oquab, Maxime and Jose, Cijo and Khalidov, Vasil and Szafraniec, Marc and Yi, Seungeun and Ramamonjisoa, Micha{\"e}l and others},
  journal={arXiv preprint arXiv:2508.10104},
  year={2025}
}

@article{dosovitskiy2020image,
  title={An image is worth 16x16 words: Transformers for image recognition at scale},
  author={Dosovitskiy, Alexey},
  journal={arXiv preprint arXiv:2010.11929},
  year={2020}
}

@article{huang2025real,
  title={Real-Time Object Detection Meets DINOv3},
  author={Huang, Shihua and Hou, Yongjie and Liu, Longfei and Yu, Xuanlong and Shen, Xi},
  journal={arXiv preprint arXiv:2509.20787},
  year={2025}
}

@inproceedings{peebles2023scalable,
  title={Scalable diffusion models with transformers},
  author={Peebles, William and Xie, Saining},
  booktitle={Proceedings of the IEEE/CVF international conference on computer vision},
  pages={4195--4205},
  year={2023}
}

@inproceedings{li2022bevformer,
  title={BEVFormer: Learning Bird's-Eye-View Representation from Multi-Camera Images via Spatiotemporal Transformers},
  author={Li, Zhiqi and Wang, Wenhai and Li, Hongyang and Xie, Enze and Sima, Chonghao and Lu, Tong and Yu, Qiao and Dai, Jifeng},
  booktitle={European Conference on Computer Vision},
  year={2022},
  eprint={2203.17270}
}

@article{zhu2020deformable,
  title={Deformable detr: Deformable transformers for end-to-end object detection},
  author={Zhu, Xizhou and Su, Weijie and Lu, Lewei and Li, Bin and Wang, Xiaogang and Dai, Jifeng},
  journal={arXiv preprint arXiv:2010.04159},
  year={2020}
}

@article{chitta2022transfuser,
  title={Transfuser: Imitation with transformer-based sensor fusion for autonomous driving},
  author={Chitta, Kashyap and Prakash, Aditya and Jaeger, Bernhard and Yu, Zehao and Renz, Katrin and Geiger, Andreas},
  journal={IEEE transactions on pattern analysis and machine intelligence},
  volume={45},
  number={11},
  pages={12878--12895},
  year={2022},
  publisher={IEEE}
}

@article{guo2025ipad,
  title={iPad: Iterative Proposal-centric End-to-End Autonomous Driving},
  author={Guo, Ke and Liu, Haochen and Wu, Xiaojun and Pan, Jia and Lv, Chen},
  journal={arXiv preprint arXiv:2505.15111},
  year={2025}
}

@inproceedings{gupta2018social,
  title={Social gan: Socially acceptable trajectories with generative adversarial networks},
  author={Gupta, Agrim and Johnson, Justin and Fei-Fei, Li and Savarese, Silvio and Alahi, Alexandre},
  booktitle={Proceedings of the IEEE conference on computer vision and pattern recognition},
  pages={2255--2264},
  year={2018}
}

@article{dauner2024navsim,
  title={Navsim: Data-driven non-reactive autonomous vehicle simulation and benchmarking},
  author={Dauner, Daniel and Hallgarten, Marcel and Li, Tianyu and Weng, Xinshuo and Huang, Zhiyu and Yang, Zetong and Li, Hongyang and Gilitschenski, Igor and Ivanovic, Boris and Pavone, Marco and others},
  journal={Advances in Neural Information Processing Systems},
  volume={37},
  pages={28706--28719},
  year={2024}
}

@article{chen2024vadv2,
  title={Vadv2: End-to-end vectorized autonomous driving via probabilistic planning},
  author={Chen, Shaoyu and Jiang, Bo and Gao, Hao and Liao, Bencheng and Xu, Qing and Zhang, Qian and Huang, Chang and Liu, Wenyu and Wang, Xinggang},
  journal={arXiv preprint arXiv:2402.13243},
  year={2024}
}

@article{li2024hydra,
  title={Hydra-mdp: End-to-end multimodal planning with multi-target hydra-distillation},
  author={Li, Zhenxin and Li, Kailin and Wang, Shihao and Lan, Shiyi and Yu, Zhiding and Ji, Yishen and Li, Zhiqi and Zhu, Ziyue and Kautz, Jan and Wu, Zuxuan and others},
  journal={arXiv preprint arXiv:2406.06978},
  year={2024}
}

@inproceedings{radford2021learning,
  title={Learning transferable visual models from natural language supervision},
  author={Radford, Alec and Kim, Jong Wook and Hallacy, Chris and Ramesh, Aditya and Goh, Gabriel and Agarwal, Sandhini and Sastry, Girish and Askell, Amanda and Mishkin, Pamela and Clark, Jack and others},
  booktitle={International conference on machine learning},
  pages={8748--8763},
  year={2021},
  organization={PmLR}
}

@inproceedings{jia2021scaling,
  title={Scaling up visual and vision-language representation learning with noisy text supervision},
  author={Jia, Chao and Yang, Yinfei and Xia, Ye and Chen, Yi-Ting and Parekh, Zarana and Pham, Hieu and Le, Quoc and Sung, Yun-Hsuan and Li, Zhen and Duerig, Tom},
  booktitle={International conference on machine learning},
  pages={4904--4916},
  year={2021},
  organization={PMLR}
}

@inproceedings{li2022blip,
  title={Blip: Bootstrapping language-image pre-training for unified vision-language understanding and generation},
  author={Li, Junnan and Li, Dongxu and Xiong, Caiming and Hoi, Steven},
  booktitle={International conference on machine learning},
  pages={12888--12900},
  year={2022},
  organization={PMLR}
}

@inproceedings{li2023blip,
  title={Blip-2: Bootstrapping language-image pre-training with frozen image encoders and large language models},
  author={Li, Junnan and Li, Dongxu and Savarese, Silvio and Hoi, Steven},
  booktitle={International conference on machine learning},
  pages={19730--19742},
  year={2023},
  organization={PMLR}
}

@inproceedings{huang2025deim,
  title={Deim: Detr with improved matching for fast convergence},
  author={Huang, Shihua and Lu, Zhichao and Cun, Xiaodong and Yu, Yongjun and Zhou, Xiao and Shen, Xi},
  booktitle={Proceedings of the Computer Vision and Pattern Recognition Conference},
  pages={15162--15171},
  year={2025}
}

@article{bao2021beit,
  title={Beit: Bert pre-training of image transformers},
  author={Bao, Hangbo and Dong, Li and Piao, Songhao and Wei, Furu},
  journal={arXiv preprint arXiv:2106.08254},
  year={2021}
}

@inproceedings{he2022masked,
  title={Masked autoencoders are scalable vision learners},
  author={He, Kaiming and Chen, Xinlei and Xie, Saining and Li, Yanghao and Doll{\'a}r, Piotr and Girshick, Ross},
  booktitle={Proceedings of the IEEE/CVF conference on computer vision and pattern recognition},
  pages={16000--16009},
  year={2022}
}

@inproceedings{xie2022simmim,
  title={Simmim: A simple framework for masked image modeling},
  author={Xie, Zhenda and Zhang, Zheng and Cao, Yue and Lin, Yutong and Bao, Jianmin and Yao, Zhuliang and Dai, Qi and Hu, Han},
  booktitle={Proceedings of the IEEE/CVF conference on computer vision and pattern recognition},
  pages={9653--9663},
  year={2022}
}

@inproceedings{baevski2022data2vec,
  title={Data2vec: A general framework for self-supervised learning in speech, vision and language},
  author={Baevski, Alexei and Hsu, Wei-Ning and Xu, Qiantong and Babu, Arun and Gu, Jiatao and Auli, Michael},
  booktitle={International conference on machine learning},
  pages={1298--1312},
  year={2022},
  organization={PMLR}
}

@inproceedings{xia2022vision,
  title={Vision transformer with deformable attention},
  author={Xia, Zhuofan and Pan, Xuran and Song, Shiji and Li, Li Erran and Huang, Gao},
  booktitle={Proceedings of the IEEE/CVF conference on computer vision and pattern recognition},
  pages={4794--4803},
  year={2022}
}

@article{liu2022dab,
  title={Dab-detr: Dynamic anchor boxes are better queries for detr},
  author={Liu, Shilong and Li, Feng and Zhang, Hao and Yang, Xiao and Qi, Xianbiao and Su, Hang and Zhu, Jun and Zhang, Lei},
  journal={arXiv preprint arXiv:2201.12329},
  year={2022}
}

@inproceedings{meng2021conditional,
  title={Conditional detr for fast training convergence},
  author={Meng, Depu and Chen, Xiaokang and Fan, Zejia and Zeng, Gang and Li, Houqiang and Yuan, Yuhui and Sun, Lei and Wang, Jingdong},
  booktitle={Proceedings of the IEEE/CVF international conference on computer vision},
  pages={3651--3660},
  year={2021}
}

@inproceedings{wang2022detr3d,
  title={Detr3d: 3d object detection from multi-view images via 3d-to-2d queries},
  author={Wang, Yue and Guizilini, Vitor Campagnolo and Zhang, Tianyuan and Wang, Yilun and Zhao, Hang and Solomon, Justin},
  booktitle={Conference on robot learning},
  pages={180--191},
  year={2022},
  organization={PMLR}
}

@inproceedings{girshick2014rich,
  title={Rich feature hierarchies for accurate object detection and semantic segmentation},
  author={Girshick, Ross and Donahue, Jeff and Darrell, Trevor and Malik, Jitendra},
  booktitle={Proceedings of the IEEE conference on computer vision and pattern recognition},
  pages={580--587},
  year={2014}
}

@article{ren2015faster,
  title={Faster r-cnn: Towards real-time object detection with region proposal networks},
  author={Ren, Shaoqing and He, Kaiming and Girshick, Ross and Sun, Jian},
  journal={Advances in neural information processing systems},
  volume={28},
  year={2015}
}

@inproceedings{jia2023think,
  title={Think twice before driving: Towards scalable decoders for end-to-end autonomous driving},
  author={Jia, Xiaosong and Wu, Penghao and Chen, Li and Xie, Jiangwei and He, Conghui and Yan, Junchi and Li, Hongyang},
  booktitle={Proceedings of the IEEE/CVF Conference on Computer Vision and Pattern Recognition},
  pages={21983--21994},
  year={2023}
}

@inproceedings{wang2024trajfine,
  title={TrajFine: Predicted trajectory refinement for pedestrian trajectory forecasting},
  author={Wang, Kuan-Lin and Tsao, Li-Wu and Wu, Jhih-Ciang and Shuai, Hong-Han and Cheng, Wen-Huang},
  booktitle={Proceedings of the IEEE/CVF Conference on Computer Vision and Pattern Recognition},
  pages={4483--4492},
  year={2024}
}

@article{yin2025diffrefiner,
  title={DiffRefiner: Coarse to Fine Trajectory Planning via Diffusion Refinement with Semantic Interaction for End to End Autonomous Driving},
  author={Yin, Liuhan and Ju, Runkun and Guo, Guodong and Cheng, Erkang},
  journal={arXiv preprint arXiv:2511.17150},
  year={2025}
}

@article{bojarski2016end,
  title={End to end learning for self-driving cars},
  author={Bojarski, Mariusz and Del Testa, Davide and Dworakowski, Daniel and Firner, Bernhard and Flepp, Beat and Goyal, Prasoon and Jackel, Lawrence D and Monfort, Mathew and Muller, Urs and Zhang, Jiakai and others},
  journal={arXiv preprint arXiv:1604.07316},
  year={2016}
}

@inproceedings{codevilla2018end,
  title={End-to-end driving via conditional imitation learning},
  author={Codevilla, Felipe and M{\"u}ller, Matthias and L{\'o}pez, Antonio and Koltun, Vladlen and Dosovitskiy, Alexey},
  booktitle={2018 IEEE international conference on robotics and automation (ICRA)},
  pages={4693--4700},
  year={2018},
  organization={IEEE}
}

@article{huang2021bevdet,
  title={Bevdet: High-performance multi-camera 3d object detection in bird-eye-view},
  author={Huang, Junjie and Huang, Guan and Zhu, Zheng and Ye, Yun and Du, Dalong},
  journal={arXiv preprint arXiv:2112.11790},
  year={2021}
}

@article{huang2022bevdet4d,
  title={Bevdet4d: Exploit temporal cues in multi-camera 3d object detection},
  author={Huang, Junjie and Huang, Guan},
  journal={arXiv preprint arXiv:2203.17054},
  year={2022}
}

@article{wu2022yolop,
  title={Yolop: You only look once for panoptic driving perception},
  author={Wu, Dong and Liao, Man-Wen and Zhang, Wei-Tian and Wang, Xing-Gang and Bai, Xiang and Cheng, Wen-Qing and Liu, Wen-Yu},
  journal={Machine Intelligence Research},
  volume={19},
  number={6},
  pages={550--562},
  year={2022},
  publisher={Springer}
}

@inproceedings{hu2023planning,
  title={Planning-oriented autonomous driving},
  author={Hu, Yihan and Yang, Jiazhi and Chen, Li and Li, Keyu and Sima, Chonghao and Zhu, Xizhou and Chai, Siqi and Du, Senyao and Lin, Tianwei and Wang, Wenhai and others},
  booktitle={Proceedings of the IEEE/CVF conference on computer vision and pattern recognition},
  pages={17853--17862},
  year={2023}
}

@inproceedings{jiang2023vad,
  title={Vad: Vectorized scene representation for efficient autonomous driving},
  author={Jiang, Bo and Chen, Shaoyu and Xu, Qing and Liao, Bencheng and Chen, Jiajie and Zhou, Helong and Zhang, Qian and Liu, Wenyu and Huang, Chang and Wang, Xinggang},
  booktitle={Proceedings of the IEEE/CVF International Conference on Computer Vision},
  pages={8340--8350},
  year={2023}
}

@article{yuan2024drama,
  title={Drama: An efficient end-to-end motion planner for autonomous driving with mamba},
  author={Yuan, Chengran and Zhang, Zhanqi and Sun, Jiawei and Sun, Shuo and Huang, Zefan and Lee, Christina Dao Wen and Li, Dongen and Han, Yuhang and Wong, Anthony and Tee, Keng Peng and others},
  journal={arXiv preprint arXiv:2408.03601},
  year={2024}
}

@inproceedings{weng2024drive,
  title={Para-drive: Parallelized architecture for real-time autonomous driving},
  author={Weng, Xinshuo and Ivanovic, Boris and Wang, Yan and Wang, Yue and Pavone, Marco},
  booktitle={Proceedings of the IEEE/CVF Conference on Computer Vision and Pattern Recognition},
  pages={15449--15458},
  year={2024}
}
